\RequirePackage{letltxmacro}
\LetLtxMacro{\LaTeXtextbf}{\textbf}
\documentclass{ieeeaccess}
\LetLtxMacro{\textbf}{\LaTeXtextbf}

\usepackage{cite}
\usepackage{amsmath,amssymb,amsfonts}
\usepackage{algorithmic}
\usepackage{graphicx}
\usepackage{textcomp}
\usepackage{hyperref}

\usepackage[linesnumbered,ruled]{algorithm2e}

\def\BibTeX{{\rm B\kern-.05em{\sc i\kern-.025em b}\kern-.08em
    T\kern-.1667em\lower.7ex\hbox{E}\kern-.125emX}}
\begin{document}
\history{}
\doi{}

\title{A Unifying Framework of Attention-based Neural Load Forecasting}
\author{\uppercase{Jing Xiong} and
\uppercase{Yu Zhang}}
\address[]{Department of Electrical and Computer Engineering, University of California Santa Cruz, CA 95064 USA}

\tfootnote{This work was supported in part by the Hellman Fellowship, and a Seed Fund Award from CITRIS and the Banatao Institute at the University of California.}

\markboth
{Xiong \headeretal: A Unifying Framework of Attention-based Neural Load Forecasting}
{Xiong \headeretal: A Unifying Framework of Attention-based Neural Load Forecasting}

\corresp{Corresponding author: Yu Zhang (Email: yzhan419@ucsc.edu).}

\begin{abstract}
Accurate load forecasting is critical for reliable and efficient planning and operation of electric power grids. In this paper, we propose a unifying deep learning framework for load forecasting, which includes time-varying feature weighting, hierarchical temporal attention, and feature-reinforced error correction. Our framework adopts a modular design with good generalization capability. First, the feature-weighting mechanism assigns input features with temporal weights. Second, a recurrent encoder-decoder structure with hierarchical attention is developed as a load predictor. The hierarchical attention enables a similar day selection, which re-evaluates the importance of historical information at each time step. Third, we develop an error correction module that explores the errors and learned feature hidden information to further improve the model's forecasting performance. Experimental results demonstrate that our proposed framework outperforms existing methods on two public datasets and performance metrics, with the feature weighting mechanism and error correction module being critical to achieving superior performance. Our framework provides an effective solution to the electric load forecasting problem, which can be further adapted to many other forecasting tasks.
\end{abstract}

\begin{keywords}
Short-term load forecasting, feature weighting, attention mechanism, error correction.
\end{keywords}

\titlepgskip=-15pt
\maketitle

\section{Introduction}
\label{sec:introduction}
\PARstart{L}{oad} forecasting refers to the prediction of the future load behavior, which can be derived from the historical load pattern and its relevant features. Based on different forecast horizons, load forecasting can be divided into three categories short-term (one hour to a week), medium-term (one week to a year), and long-term (one to twenty years) forecasting. 
Each of them benefits various applications and business needs. Long-term forecasting is mainly used in power system planning such as generation and transmission expansion planning. Medium-term forecasting plays a crucial role in maintenance scheduling. Short-term load forecasting (STLF) is indispensable for day-ahead unit commitment, market clearing, spinning reserve plans, energy bidding, as well as economic load dispatch \cite{feng2019reinforced}. 

As the forecasting time span shrinks, the requirement for forecasting accuracy increases. In addition, wide applications of renewable energy generations, energy storage systems, and electric vehicles in recent years have had a huge impact on users' load behavior. This poses significant challenges in forecasting load demand \cite{kong2020short}. Various approaches have been proposed to improve the STLF accuracy. They can be roughly categorized into three classes: (i) time series analysis, (ii) classical machine learning algorithms, and (iii) deep learning models. 

Time series analysis has been widely used in many applications. Various versions of autoregressive integrated moving average (ARIMA) models were used for STLF \cite{juberias1999new}. These methods are easily implemented and interpreted. However, they often require meticulous preprocessing to make a time series stationary \cite{contreras2003arima}. Moreover, time series approaches are sensitive to irrelevant features and may fail to capture a long-term dependency.

With the vigorous development of classical machine learning theory, researchers began to explore its application in STLF. Ceperic \textit{et al.} proposed a support vector regression machines (SVR) approach for STLF, which minimizes the user interaction requirement by an adaptive model building strategy \cite{ceperic2013strategy}. A random forest (RF) model was used to deal with non-stationarity, heteroscedasticity, trend, and multiple seasonal cycles in load data \cite{dudek2015short}. In order to avoid information loss, Cheng \textit{et al.} used different feature sets to construct an ensemble random forest-based model \cite{cheng2012random}. Taieb \textit{et al.} implemented component-wise gradient boosting models (GBM) for each hour for multi-step STLF \cite{taieb2014gradient}. These classic machine learning models, which are more robust to tolerate irrelevant features, are capable of capturing nonlinear behaviors of electricity load. However, most of them use predetermined nonlinear models, which may prevent them from effectively learning the true underlying mappings \cite{qin2017dual}.

In the last decade, deep neural networks have demonstrated remarkable capabilities in uncovering complex input-output relationships in various fields, such as natural language processing and computer vision. The Deep Belief Network (DBN) is a prevalent model for time-series forecasting tasks. To improve the forecasting accuracy, rough set theory was introduced in \cite{khodayar2017rough} to enhance the feature extraction capability of the restricted Boltzmann machine (RBM) within the DBN. Furthermore, interval probability distribution learning (IPDL) in \cite{khodayar2018interval} uses deep generative neural networks to learn the input data distribution and provide uncertainty intervals for wind speed forecasting. Recurrent neural networks (RNNs) are commonly used with proven efficacy for sequence-to-sequence (seq2seq) learning and time-series forecasting tasks. However, the vanilla RNNs suffer from the gradient vanishing issue that limits their performance. To address this issue, Hochreiter and Schmidhuber proposed long short-term memory networks (LSTM), which use a gating technique to control information flows. LSTM uses three gates (input/forget/output) to retain relevant information for long-term memory while discarding the other information \cite{hochreiter1997long}. In 2014, Cho \textit{et al.} proposed gated recurrent units (GRUs) which is another gating mechanism-based RNN. GRUs reduce the number of gates with fewer parameters to train; see details in \cite{cho2014learning}.

Seq2seq learning solves the mapping between the sequential inputs and outputs of the task, which shares various similarities with time-series learning problems \cite{sutskever2014sequence}. The encoder-decoder structure usually serves as the backbone for most seq2seq models \cite{vaswani2017attention,cho2014properties}. Specifically, in time-series tasks, the encoder encodes the historical input feature sequence into a single fixed-length vector based on which the decoder yields the output. However, coping with long input sequences can be challenging. To mitigate this drawback, the attention mechanism was introduced to search for a set of positions in historical time steps where the most relevant information can be concentrated \cite{bahdanau2014neural}. For this new paradigm, a context vector is designed to bridge the encoder and the decoder, which is filtered for each output time step.

It is often challenging to deal with various conditions in real data by using single-module approaches. To further improve the prediction accuracy, hybrid models combining the advantages of all added modules have been developed. When it comes to the STLF, feature engineering \cite{kwak2002input} and error correction modules play an important role. Traditional feature selection approaches, such as filters, wrappers and embedding methods, aim at selecting the smallest subset of features that contribute the most to the output.  A two-stage hybrid model for STLF is proposed in \cite{ghadimi2018two}. Based on the mutual information criterion, the selected features are fed into a forecast engine that is implemented via Ridgelet and Elman neural networks.

Rather than selecting a subset of features, feature weighting attempts to weight each feature based on their importance or relevance with the output \cite{panday2018feature}. 
Utilizing the feature weights given by the random forest, Xuan proposed a multi-model fusion based deep neural network to forecast the load demand \cite{xuan2021multi}.
Qin \textit{et al.} proposed an input attention layer as feature weighting that can be trained simultaneously with the model \cite{qin2017dual}. However, their scheme is based only on the past information which cannot capture all the information of the entire input sequence. Moreover, the feature weighting part is embedded in the encoder, which makes it hard to be adopted for other basic structures; e.g. the convolutional neural network (CNN).

The prediction error generally comes from two parts: the learning capability of the original model and the newly emerging unknown data. To further improve the prediction accuracy, error correction module can learn useful hidden information from the error values. In this context, Deng \textit{et al.} proposed a hybrid model which includes a decomposition module, a forecasting module, and an error correction module for wind speed forecasting \cite{deng2020hybrid}. Leveraging the dynamic mode decomposition (DMD) method in fluid dynamics, Kong \textit{et al.} captured the spatio-temporal dynamics of error series in STLF \cite{kong2020short}. This algorithm first constructs the error Hankel matrix and then does the pattern decomposition of the error. Existing approaches for the error correction are based on either a completely new model such as ARIMA \cite{duan2021short}, \cite{deng2020hybrid} or extreme learning machine \cite{liu2019multi}. However, the design of a new model will increase the learning cost. Model selection and hyperparameter tuning are necessary, which may greatly affect the sampling complexity and training time. In addition, the useful knowledge learned by the predictive model will be lost in the new model. 

Transfer learning was proposed to deal with the aforementioned issues. The motivation is to use previously acquired domain knowledge to solve new problems faster, or yield better solutions \cite{pan2009survey}. In recent years, transfer learning has been successfully used for supervised and unsupervised learning. In load forecasting, researchers have also explored this technique, where the knowledge is transferred from one region/household to another one \cite{cai2019two}. In this case, the source and target domains are the same, which is load and relative features while the task is also the same. The key challenge for applying transfer learning in error correction is incorporating the error information into the target domain without changing the input dimension while capitalizing on previously acquired feature knowledge. 

Considering the limitations of the aforementioned prior works, in this paper we propose a novel deep learning framework that incorporates a dynamic feature weighting mechanism and a transferred learning based error correction module. The main contributions of our work are listed as follows:
\begin{enumerate}
\item The proposed framework offers a modular and plug-and-play functionality that can be adapted to different types of data and setups in STLF.
\item  Compared with classic feature weighting methods, our attention-based time-varying feature weighting mechanism can assign different levels of importance to each feature at each time step, which allows for more dynamic and adaptive feature selection.
\item The hierarchical temporal attention layer captures the similar day and similar hour information, which is a critical factor for accurate STLF.
\item The proposed error correction module leverages transfer learning. This eliminates the need for a new model design and inherits the learned feature knowledge.
\item Extensive experimental results corroborate the merits of our approach, which outperforms existing methods based on various performance metrics. 
\end{enumerate}

The remainder of the paper is organized as follows. Section II presents the overall framework with all proposed modules. The simulation setup and results are reported in Section III and IV, respectively. Finally, Section V summarizes this work.

\section{The Proposed Load Forecasting Framework}

\begin{figure}[t]
\centering
\includegraphics[width=0.48\textwidth]{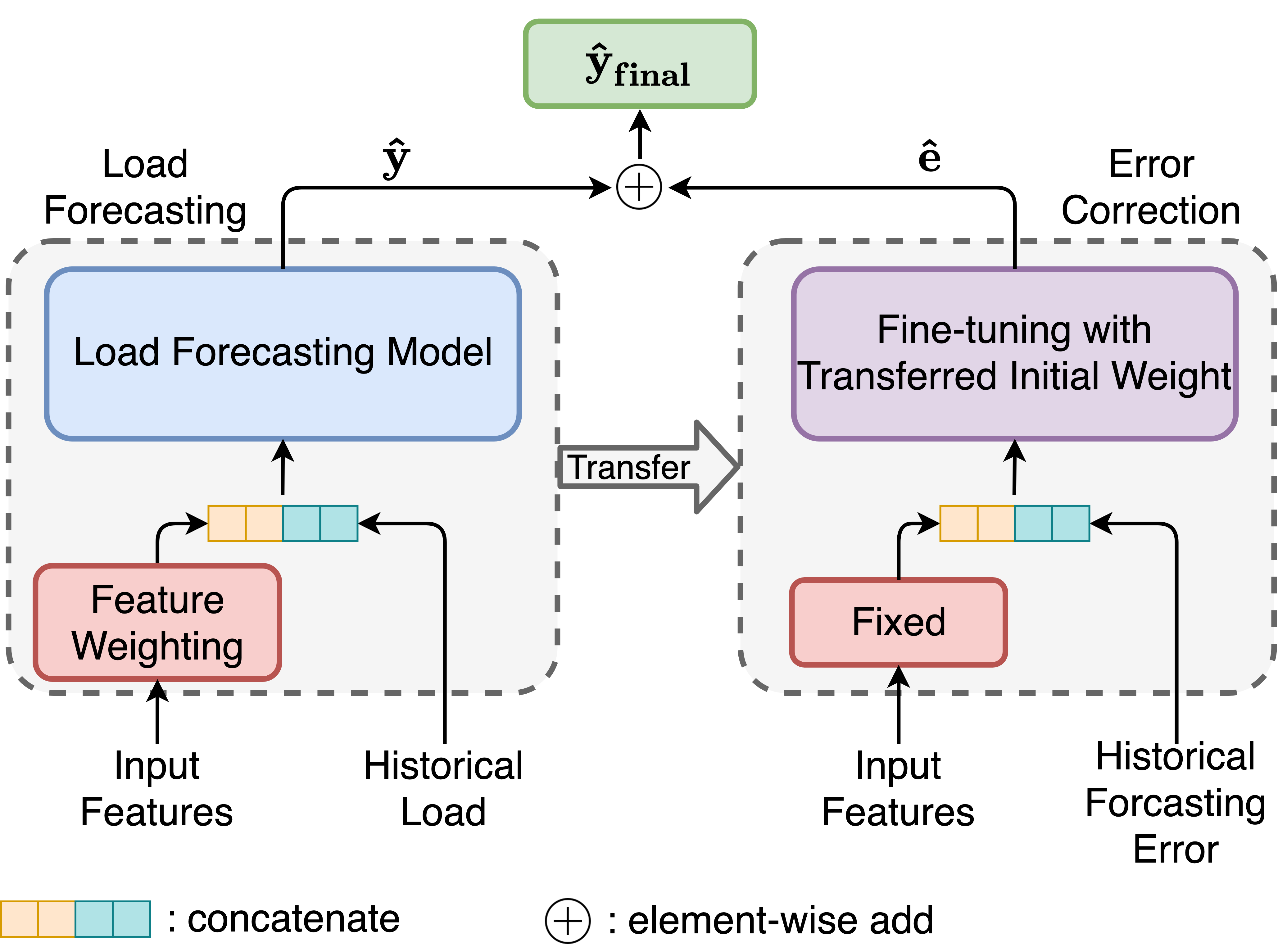}
\caption{The architecture of the proposed framework. The left side is the load forecasting module with an input feature-weighting mechanism designed to weigh the different input features. The right side is the error correction module transferred from the left side model to further enhance forecasting ability. The detailed structures of feature-weighting mechanism and load forecasting module are shown in Fig.~\ref{fig:FW} and Fig.~\ref{fig:ForecastingModel}.}\label{fig:overall_structure}
\vspace{-0.2cm}
\end{figure}

In the following section, we introduce the overall load forecasting framework structure as shown in Fig.~\ref{fig:overall_structure}. The framework mainly consists of three modules: (i) the feature-weighting mechanism, (ii) the short-term load forecasting module, and (iii) the error correction module. 

\subsection{Feature Embedding and Feature-weighting Mechanism}

Input features can generally be divided into two categories: numeric features and categorical features. As the model requires numeric input, a categorical feature would be transformed into a numeric vector. 
For STLF, the inputs can contain meteorological conditions (e.g., temperature, humidity, wind speed and direction, etc), time-related features (e.g., indicators of holidays, seasons, etc), and utility discount programs. For those categorical features, we use one-hot encoding in this work. After the embedding, all input features $\mathbf{X}=(\mathbf{x}^1,\mathbf{x}^2,\dots,\mathbf{x}^n) \in \mathbb{R}^{(T_h+T_f)\times n}$  are the concatenation of encoded categorical features and continuous numeric features. Each row of $\mathbf{X}$, denoted as $\mathbf{x}_{t} = (x_{t}^1,x_{t}^2,\dots,x_{t}^n)$, represents all features at time $t$. 

Feature selection plays a crucial role in machine learning methods \cite{kwak2002input}. Irrelevant features can significantly affect the model's performance. Instead of making a hard feature selection which is a special case of feature weighting mechanism, the proposed model is able to adaptively weigh different features and give more attention to features that contribute more to the target values. 
In \cite{xiong2021short}, the feature selection layer is entangled in the encoder. Therefore, it is hard to transfer to other load forecasting modules. In order to modularize the proposed framework, we separate the feature selection layer from the encoder, and the weight $\alpha_t^k$ of each feature $k$ at time $t$ is calculated via the softmax operator:
\begin{equation} \label{eq:weightAlpha}
    \alpha_{t}^{k}=\frac{\exp \left(h_{t}^{k}\right)}{\sum_{i=1}^{n} \exp \left(h_{t}^{i}\right)},\, k = 1,2,\ldots, n,
\end{equation}
where $h_t^k$ is the $k$-th entry of the vector 
$
\mathbf{h}_{t}=\mathbf{V}_{\alpha}\tanh\left(\mathbf{W}_{\alpha}\mathbf{x}_{t}^{\top}\right) \in \mathbb{R}^n.
$

\begin{figure}[t]
\centering
\includegraphics[width=0.48\textwidth]{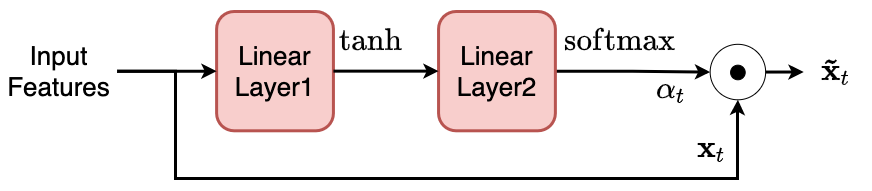}
\caption{The feature-weighting mechanism structure with two linear layers.}
\label{fig:FW}
\vspace{-0.2cm}
\end{figure}

The weight matrices $\mathbf{V}_{\alpha} \in \mathbb{R}^{n\times d^{fw}_h}$ and $\mathbf{W}_{\alpha} \in \mathbb{R}^{d^{fw}_h\times n}$ are trained jointly with the proposed model. $d^{fw}_h$ is the number of neurons in the hidden layer and it's a hyper-parameter to tune. We omit the bias term for succinctness. Then, the weighted feature input is given by 
$\mathbf{\tilde{x}}_{t} =\mathbf{\alpha}_{t} \odot \mathbf{x}_{t}$,
where $\odot$ denotes the element-wise multiplication. The detailed structure for the feature weighting mechanism is shown in Fig.~\ref{fig:FW}. 

The encoder's inputs are the concatenation of embedded categorical features, continuous numeric features, and historical target values (active power demand) at each time step $t\in [t-T_h+1,t]$. The decoder's inputs are the concatenation of embedded categorical features and continuous numeric features at each time step $t\in [t+1,t+T_f]$. $T_h$ and $T_f$ are the window size of historical and future data, respectively.

\subsection{Short-term Load Forecasting Model}

\subsubsection{Encoder-decoder structure}
The encoder-decoder structure is a workhorse in state-of-the-art deep neural networks. For time series forecasting, the encoder maps historical input features $\mathbf{x}_{t}$ and output $y_{h,t}$ at each time step to a hidden vector $\mathbf{h}_t$ that is passed to the decoder. Then, the decoder uses the last step hidden state of the encoder as its initial hidden state, and outputs future target values based on future feature inputs. In this work, a bi-directional recurrent layer (RL) is used for both encoder and decoder. We abbreviate the formulation for bi-directional RL as $\mathbf{h}_{t}=\mathrm{BiRL}(\cdot)$, where RL can be chosen as recurrent neural network (RNN), LSTM or gated recurrent unit (GRU).

\begin{figure}[t]
\centering
\includegraphics[width=0.48\textwidth]{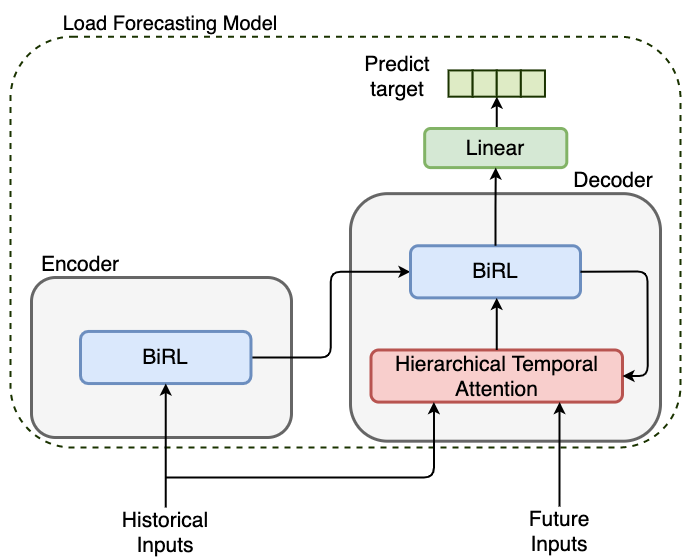}
\caption{The network architecture of the load forecasting model. The hierarchical temporal attention in the decoder focuses on the temporal similarity to incorporate similar day information.}
\label{fig:ForecastingModel}
\end{figure}

The BiRL consists of two sub-layers in opposite directions which can capture the complete information of the entire input sequence at each time step. Let $ \mathbf{h}_{t}^f$,  $ \mathbf{h}_{t}^b \in \mathbb{R}^{hs}$ denote the hidden state of forward and backward recurrent layer at time $t$, respectively. Given a sequence of historical weighted feature and target value pairs $(\tilde{\mathbf{x}}_t, y_t)$, the encoder's hidden states are updated from time $t-T_h+1$ to $t$ as
\begin{subequations} 
\label{eq:biLSTM}
\begin{align}
    \mathbf{h}_{t}^f &=\mathrm{RL}^f(\mathbf{h}_{t-1}^f,[\tilde{\mathbf{x}}_t;y_t]^{\top}),\vspace{1ex}\\
    \mathbf{h}_{t}^b &=\mathrm{RL}^b(\mathbf{h}_{t+1}^b,[\tilde{\mathbf{x}}_t;y_t]^{\top}),\vspace{1ex}\\    
    \mathbf{h}_{t}^e &=\left[{\mathbf{h}_{t}^f}^{\top};{\mathbf{h}_{t}^b}^{\top}\right]^{\top}\vspace{1ex},
\end{align}
\end{subequations}
where $[\mathbf{a}; \mathbf{b}]$ denotes the concatenation of vectors $\mathbf{a}$ and $\mathbf{b}$.

Using future weighted feature $\tilde{\mathbf{x}}_t$ and context vector of hierarchical temporal attention $\mathbf{a}_t$ (see next subsection) as inputs,
the decoder updates the hidden state iteratively from time $t+1$ to $ t+T_f$ with  
initial state $\mathbf{h}_{t}^e$. Hence, we have $\mathbf{h}_{t}^d=\mathrm{BiRL}(\mathbf{h}_{t}^e,[\tilde{\mathbf{x}}_t;\mathbf{a}_t^{\top}])$ with the detailed steps as
\begin{subequations} 
\label{eq:biLSTMdecoder}
\begin{align}
    \mathbf{h}_{t}^f &=\mathrm{RL}^f(\mathbf{h}_{t-1}^f,[\tilde{\mathbf{x}}_t;\mathbf{a}_t^{\top}]),\vspace{1ex}\\
    \mathbf{h}_{t}^b &=\mathrm{RL}^b(\mathbf{h}_{t+1}^b,[\tilde{\mathbf{x}}_t;\mathbf{a}_t^{\top}]),\vspace{1ex}\\    
    \mathbf{h}_{t}^d &=\left[{\mathbf{h}_{t}^f}^{\top};{\mathbf{h}_{t}^b}^{\top}\right]^{\top}\vspace{1ex}.
\end{align}
\end{subequations}

Finally, a fully connected layer with the rectified linear unit (ReLU) activation function is used to transform the hidden information to the forecast output from time $t+1$ to $ t+T_f$:
\begin{equation} \label{eq:output}
    \mathbf{y}_{f,t}=\mathbf{V}_y\text{ReLU}(\mathbf{W}_y\mathbf{h}_t^d),
\end{equation}
where $\mathbf{V}_y \in \mathbb{R}^{1 \times d^{o}_h}$ and $\mathbf{W}_y \in \mathbb{R}^{d^{o}_h\times 2hs}$ are weight matrices.

\subsubsection{Hierarchical temporal attention mechanism}

Incorporating the information of similar days and hours has been considered in the literature for load forecasting; see e.g., \cite{feng2018hourly, barman2018regional}. However, such information is often treated as additional input features or used to generate separate models. This paper uses a novel hierarchical temporal attention layer designed from our previous work \cite{xiong2021short}, which incorporates a similar day soft selection to re-evaluate the importance of historical information at each time step $t$. 

Consider using previous $M$ days of historical data to forecast the hourly loads for the next day, where each day includes $t_d =24$ data points. Thus, we have $T_h = M\times t_d$ and $T_f = t_d$. 
Let $\mathbf{X}_i=(\mathbf{x}_i^1,\mathbf{x}_i^2,\dots,\mathbf{x}_i^n)\in \mathbb{R}^{t_d\times n}$ and $\mathbf{X}_f=(\mathbf{x}_f^1,\mathbf{x}_f^2,\dots,\mathbf{x}_f^n)\in \mathbb{R}^{t_d\times n}$ collect the historical features for the day $i$ and the future features of the next day. We use the sum of feature-by-feature dissimilarities $D(\mathbf{X}_i,\mathbf{X}_f) = \sum_{k=1}^n \|\mathbf{x}_i^k-\mathbf{x}_f^k\|_2$ to quantify the distance between all features of those two days. Then, the similar day weight $\gamma_{i}$ is calculated as the softmax of the reciprocal of the distance:
\begin{equation} \label{eq:weightGamma}
    \gamma_{i}=\frac{\exp\left(D^{-1}(\mathbf{X}_i,\mathbf{X}_f)\right)}{\sum_{i=1}^M\exp\left(D^{-1}(\mathbf{X}_i,\mathbf{X}_f)\right)},\, i=1,2,\ldots,M.
\end{equation}

When forecasting load at time $t$, not all historical data contribute equally to the model's output. Hence, the attention mechanism facilitates the extraction of historical information that is more important to the current forecast value. Let subscript $i$ denote the $i$-th day and $j$ for $j$-th hour. Then, the attention weight $\beta_{i,j,t}$ is given by
\begin{equation} \label{eq:weightBeta}
    \beta_{i,j,t}=\frac{\exp(d_{i,j,t})}{\sum_{i=1}^{M}\sum_{j=1}^{t_d}\exp(d_{i,j,t})},
\end{equation}
where $d_{i,j,t}$ is the $(i\times t_d + j)$-th element of vector $\mathbf{d}_t = (d_t^1,d_t^2,\dots,d_t^{T_h})^{\top} \in \mathbb{R}^{T_h}$, which is given as
\begin{equation} \label{eq:temporalAttention_d}
    \mathbf{d}_{t}=\mathbf{V}_d\tanh\left(\mathbf{W}_{d}\left[{\mathbf{h}_{t-1}^d}^{\top} ; \mathbf{x}_t\right]^{\top}\right).
\end{equation}
The two weight matrices $\mathbf{V}_d \in \mathbb{R}^{T_h\times d^{att}_h}$ and $\mathbf{W}_{d} \in \mathbb{R}^{d^{att}_h\times (2hs+n)}$ are trained jointly with the proposed model. 
$\mathbf{h}_{t-1}^d$ is the hidden vector of the decoder BiLSTM at time $t-1$. 

To this end, let $\mathbf{h}_{i,j}$ denote the historical hidden state for the $j$-th hour in the $i$-th day  from the encoder. The context vector of hierarchical temporal attention is calculated as
$\mathbf{a}_{t}=\sum_{i=1}^{M}\sum_{j=1}^{t_d} \gamma_{i}  \beta_{i,j,t}\mathbf{h}_{i,j}
$.

\begin{algorithm}[!bt]
\caption{Training Procedure of the Framework}
\label{alg:training}
\SetKwInOut{Input}{Input}
\SetKwInOut{Output}{Output}

\Input{Forecasting module training set $\mathcal{D}_l = \{\mathbf{X}, \mathbf{y}_h, \mathbf{y}_f\}$, \\
Error correction module training set \\
$\mathcal{D}_e = \{\mathbf{X}_{e}, \mathbf{y}_{e,h}, \mathbf{y}_{e,f}\}$, \\
Number of epochs $N_l$ and $N_e$,\\
Feature weighting module $f_l$, \\
Load forecasting module $g_l$}
\Output{Trained model $\{f_l, g_l, g_e\}$}
Initialize model parameters\;
\For {epoch = 1 to $N_l$}{
 \For {batch of $\{\mathbf{X}, \mathbf{y}_h, \mathbf{y}_f\} \in \mathcal{D}_l$}{
 $\tilde{{\mathbf{X}}} \gets f_l(\mathbf{X})$\;
 $\hat{\mathbf{y}}_f \gets g_l(\tilde{\mathbf{X}}, \mathbf{y}_h)$\;
 Compute training loss $L_{\mathrm{LF}}$\;
 Compute the gradient of loss\;
 Update parameters in $g_l(\cdot)$ and $f_l(\cdot)$\;
 }}
 
 $\hat{\mathbf{y}}_{e,f} \gets g_l(f_l(\mathbf{X}_{e}), \mathbf{y}_{e,h})$\;
 $\mathbf{e} \gets \mathbf{y}_{e,f} - \hat{\mathbf{y}}_{e,f}$\;
Set feature weighting module $f_e(\cdot) = f_l(\cdot)$\;
Set error correction module  $g_e(\cdot) = g_l(\cdot)$\;
Fixing the feature weighting layer, and train all other layers in the error correction module as follows: \\
\For {epoch = 1 to $N_e$}{
\For {batch of $\{\mathbf{X}_e, \mathbf{e}_h, \mathbf{e}_f\}$}{
 $\tilde{{\mathbf{X}}} \gets f_l(\mathbf{X})$\;
 $\hat{\mathbf{e}}_f \gets g_e(\tilde{\mathbf{X}}, \mathbf{e}_h)$\;
 Compute training loss $L_{\mathrm{EC}}$\;
 Compute the gradient of loss\;
 Update model parameter of $g_e(\cdot)$;
 }
}
Return trained model $\{f_l, g_l, g_e\}$.
\end{algorithm}

\subsection{Error Correction Module} 
Traditional error correction systems often involve creating a new model to forecast errors, resulting in higher learning costs and the potential loss of learnt  knowledge obtained by the original predictive model. To overcome these shortcomings, a transfer-learning-based error correction module is proposed. Transfer learning utilizes previously acquired domain knowledge to solve new problems more efficiently yielding better results. In recent years, transfer learning has succeeded in supervised and unsupervised learning, including load forecasting, where knowledge is transferred between regions or households. However, transferring knowledge in error correction requires incorporating the error information into the target domain without changing the input dimension while leveraging previously learned feature knowledge. The proposed error correction module addresses this challenge and aims to improve prediction accuracy by extracting valuable information from error values with the help of learned hidden features.

The error correction module is trained after the load forecasting module, which is first trained on dataset $\mathcal{D}_l$. Then, based on the error correction dataset $\mathcal{D}_e$, we compute the forecasting error as $\mathbf{e}=\mathbf{y}_{e,f} - \hat{\mathbf{y}}_{e,f}$, where $\mathbf{y}_{e,f}$ is the real output value and $\hat{\mathbf{y}}_{e,f}$ is the predicted value obtained by the forecasting model. The feature weighting module and error correction module are initialized by the forecasting model, with the feature weighting layer fixed and the other layers to be trained. To train the error correction module, a new dataset with feature input and forecasting error is generated and randomly split into training and validation sets. The algorithm computes the training loss and its gradient to update the error correction module via backpropagation. Upon completing the training of the error correction module, it can be used to correct forecast errors and improve forecasting accuracy. The final output is obtained as $\bar{\mathbf{y}}=\hat{\mathbf{y}}+\hat{\mathbf{e}}$, as shown in Fig.~\ref{fig:overall_structure}. The overall training procedure of the framework is summarized in Algorithm \ref{alg:training}. The proposed transfer learning based model has several advantages including no need for hyper-parameter tuning and the ability to train with limited data. It also reuses existing knowledge learned by the original model, which results in a faster learning rate.

\subsection{Loss Function}
For the load forecasting module, we choose the mean squared error (MSE) loss and introduce $\ell_1$ regularizer to encourage sparsity. The formulation is given as:
\begin{equation} \label{eq:Loss}
    L_{\mathrm{LF}} = \frac{1}{N} \sum_{i=1}^{N}(y_i - {\hat{y}}_i)^2 + \lambda \left\Vert \boldsymbol{\alpha} \right \Vert_1,
\end{equation}
where $\lambda$ is the weighting parameter balancing the data fitting loss and the sparsity-promoting $\ell_1$ penalty. 

For the error correction module, we drop the $\ell_1$ regularization term because the feature weighting layer is fixed. Hence, the loss function for training this module becomes
\begin{equation} \label{eq:EC-Loss}
L_{\mathrm{EC}} = \frac{1}{N} \sum_{i=1}^{N}(e_i - {\hat{e}}_i)^2.
\end{equation}

\section{Experiment Setup}

\subsection{Data Description}

The proposed framework is evaluated using two public datasets: the ISO New England (ISO-NE) dataset \footnote{Available at https://www.iso-ne.com/isoexpress/web/reports/load-and-demand/-/tree/dmnd} and the North-American Utility (NAU) dataset\footnote{Available at https://class.ece.uw.edu/555/el-sharkawi/index.htm}. ISO-NE annually releases reports that provide hourly historical demand and electricity pricing data for its control area and eight load zones. This paper focuses solely on the control area dataset and ignores the price-related features. The input features include day-ahead demand, dry bulb and dew point temperatures (in Fahrenheit), and time-related features. Data from 2015 to 2017 are used to train the forecasting module while 80\% of 2018's data are randomly selected to train the error correction module. The remaining 20\% data of 2018 are used for validation. The year 2019 is reserved for testing.
The NAU dataset provides electricity load, temperature, and time information from January 1, 1985 to October 12, 1992. In our study, temperature and time-related features are considered. We use the data from 1987 to 1989 for training the forecasting model and randomly select 80\% of 1990's data to train the error correction module. The remaining 20\% of 1990 is for validation while the year 1991 is for testing.

\subsection{Data Preparation}

\begin{figure}[!tb]
\centering
  \includegraphics[width=0.48\textwidth]{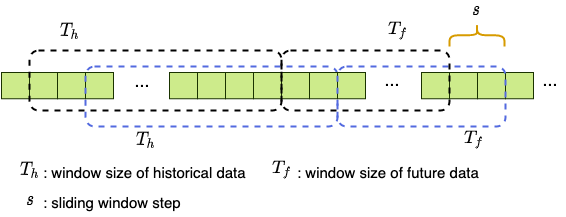}
  \caption{Illustration of the sliding window step for data processing.}\label{fig:data_split}
\end{figure}

The missing values in the datasets are filled by linear interpolation. We incorporate time-related features such as indicators of weekends and holidays, seasons, hour of day, day of week, and month of year. We embed categorical features via one-hot encoding and standardize numerical features by subtracting their means and dividing by their standard deviations. The framework forecasts next 24 hours load demands using the previous seven days load and features. The next 24 hours features are assumed to be available as model's input. For hourly data, we have $T_h=168$ and $T_f=24$. As shown in Fig.~\ref{fig:data_split}, the sliding window size is set to be 1 and 24 for training the forecasting module and error correction module, respectively. We list the model inputs from the ISO and NAU datasets in Tables \ref{table:ISO_input} and \ref{table:Utility_input}.

\begin{table}[!tb]
\centering
\caption{Time-related features of the ISO-NE dataset.}
\label{table:ISO_input}
\renewcommand{\arraystretch}{1.3}
\resizebox{0.48\textwidth}{!}
{%
\begin{tabular}{lll}
\hline
Input & Size & Description \\ \hline
$\mathbf{y}_h$ & $168\times 1$ & Historical target values  \\
DaDemd	& $192\times 1$  & Day ahead demand \\
DryBulb	& $192\times 1$  & Dry bulb temperature  \\
DewPnt & $192\times 1$  & Dew point temperature \\
Weekday & $192\times 1$  & Weekday or weekend indicator \\
Holiday  & $192\times 1$ & Holiday or non-holiday indicator \\ 
Season  & $192\times 4$ & One-hot encoding \\ 
Hour of Day  & $192\times 24$ & One-hot encoding \\
Day of Week  & $192\times 7$ & One-hot encoding   \\ 
Month of Year  & $192\times 12$ & One-hot encoding \\ \hline
\end{tabular}%
}
\end{table}

\begin{table}[!tb]
\centering
\caption{Time-related features of the NAU dataset.}
\label{table:Utility_input}
\renewcommand{\arraystretch}{1.3}
\resizebox{0.48\textwidth}{!}
{%
\begin{tabular}{lll}
\hline
Input & Size & Description \\ \hline
$\mathbf{y}_h$ & $168\times 1$ & Historical target values  \\
Temperature & $192\times 1$  & Historical and future temperature \\
Holiday  & $192\times 1$ & Holiday or non-holiday indicator \\ 
Season  & $192\times 4$ & One-hot encoding \\ 
Hour of Day  & $192\times 24$ & One-hot encoding \\
Day of Week  & $192\times 7$ & One-hot encoding   \\ 
Month of Year  & $192\times 12$ & One-hot encoding \\ \hline
\end{tabular}%
}
\end{table}

\subsection{Baseline Models and Hyperparameters}

To verify the effectiveness of our proposed framework, we compare four different types of models: classic machine learning models, DBN-based models, RNN-based models, and Transformer-based models. Details are given in below.
\begin{itemize}
\item Classic machine learning model: We test SVR \cite{ceperic2013strategy}, RF \cite{dudek2015short}, and GBM \cite{taieb2014gradient} using Scikit-Learn 0.23.2. Inputs to the model are historical and feature features and historical load that are flattened as a 1-D vector.
\item DBN-based model: DBNs \cite{dedinec2016deep} are generative neural networks composed of multiple layers of RBMs. Each RBM layer is pre-trained in an unsupervised manner using the contrastive divergence algorithm, and the overall model is fine-tuned using supervised learning. Rough autoencoder combines rough set theory with DBNs which can effectively handle uncertain and noisy data and learn complex patterns \cite{khodayar2017rough}.
\item RNN-based model: The CNN-LSTM \cite{rafi2021short} combines the advantages of both CNN and LSTM layers to improve forecasting accuracy. Attention-based load forecasting (ANLF) \cite{xiong2021short} is based on the encoder-decoder biLSTM architecture and utilizes a dynamic feature selection layer within the encoder. These models have shown promising results in load forecasting and can be used as effective baselines for future research in this field.
\item Transformer-based model: Informer is a transformer-based model designed for time-series forecasting as proposed in the 2021 AAAI Best Paper \cite{zhou2021informer}. Unlike the RNN-based model, transformer can handle sequential data in parallel to reduce training time.
\end{itemize}

The computing environment is a machine with 3.7 GHz Intel Core i7-8700K Six-Core and NVIDIA GeForce GTX 1080 Ti (11GB GDDR5X). Deep learning based models are trained by using Adam optimizer and implemented with PyTorch 1.6.0. The initial learning rate is 0.001 which decays by 0.1 times for every 30 epochs. Early stopping criteria is set with patience 30. All models share the same training, validation, and testing data samples and input features for fair comparisons. We perform a grid search to identify the best hyper-parameter set based on the validation data. The grid search is commonly used for hyper-parameter tuning, which involves setting a range of values for each hyper-parameter and testing all possible combinations. The details of the grid search are given in Table ~\ref{table:hyperparameter_tuning}.

\begin{table}[t]
\centering
\caption{The ranges of hyper-parameter tuning. Bold and italic fonts indicate the best values for the ISO-NE and the NAU datasets, respectively.}
\label{table:hyperparameter_tuning}
\renewcommand{\arraystretch}{1.25}
{%
\begin{tabular}{ll}
\hline
Model & Hyper-parameter range  \\ \hline
 SVR \cite{ceperic2013strategy} & Kernel $=$ 
 (\emph{RBF}, Linear, \textbf{Poly})\\
& Degree $=$ (2, 3) with Poly\\
& Gamma $=$ (\emph{auto}, \textbf{scale}) with Poly/RBF\\
& C $=$ (0.1, 1, \emph{\textbf{10}})\\\hline

 RF \cite{dudek2015short} & Number of estimators $=$ (100, \emph{500}, \textbf{1000})\\
&Maximum depth $=$ (5, 10, \emph{\textbf{20}} None)\\
&Minimum samples split $=$ (\emph{\textbf{2}}, 5)\\
&Minimum samples leaf $=$ (\emph{\textbf{1}}, 3, 5, 10)\\\hline

GBM \cite{taieb2014gradient} & Loss $=$ (\emph{\textbf{ls}}, lad, huber, quantile)\\
&Learning rate $=$ (\emph{0.1}, \textbf{0.01})\\
&Number of estimators $=$ (100, 500, \emph{\textbf{1000}})\\
&Maximum depth $=$ (\emph{\textbf{5}}, 10, 20, None)\\
&Minimum samples split $=$ (\emph{2}, \textbf{5})\\
&Minimum samples leaf $=$ (1, 3, \emph{5}, \textbf{10})\\\hline

DBN \cite{dedinec2016deep} & Hidden layer 1 $=$ (\emph{128}, \textbf{256}, 512)\\
&Hidden layer 2 $=$ (\emph{128}, \textbf{256}, 512)\\
&Hidden layer 3 $=$ (\emph{\textbf{0}}, 64, 128)\\
&Batch $=$ (\emph{\textbf{64}}, 128)\\\hline

RAE \cite{khodayar2017rough} & Hidden layer 1 $=$ (128, \emph{\textbf{256}}, 512)\\
&Hidden layer 2 $=$ (128, \emph{\textbf{256}}, 512)\\
&Hidden layer 3 $=$ (\emph{\textbf{0}}, 64, 128)\\
&Batch $=$ (64, \emph{\textbf{128}})\\\hline

CNN-LSTM \cite{rafi2021short} & Batch $=$ (\textbf{64}, \emph{128})\\
&Hidden size $=$ (128, 256, \emph{\textbf{512}})\\
&Kernal size $=$ (\emph{3}, 5, \textbf{8})\\\hline

ANLF\cite{xiong2021short} & Batch $=$ (64, \emph{\textbf{128}})\\
&Hidden size $=$ (\emph{\textbf{128}}, 256, 512)\\\hline

Informer\cite{zhou2021informer} & Batch $=$ (\emph{\textbf{64}}, 128)\\
&Hidden size $=$ (\textbf{64}, \emph{128}, 256, 512)\\
&label length $=$ (0, 24, \emph{\textbf{48}})\\
&number of attention heads $=$ (2, 4, \emph{\textbf{8}})\\\hline

PM-LSTM & Batch $=$ (\emph{64}, \textbf{128})\\
&Hidden size $=$ (\textbf{128}, \emph{256}, 512)\\
&$\lambda =$ (0, \emph{\textbf{0.001}}, 0.01)\\\hline

PM-GRU & Batch $=$ (64, \emph{\textbf{128}})\\
&Hidden size $=$ (\emph{128}, 256, \textbf{512})\\
&$\lambda=$ (\emph{\textbf{0}}, 0.001, 0.01)\\\hline
\end{tabular}%
}
\end{table}

\subsection{Performance Metrics}

The mean absolute error (MAE) and mean absolute percentage error (MAPE) are used to evaluate the forecasting accuracy. They are defined as follows:
\begin{subequations} 
\label{eq:evaluation}
\begin{align}
    \mathrm{MAE} &= \frac{1}{n}\sum_{i=1}^n \left|y_i - \hat{y}_i\right|,\\
    \mathrm{MAPE} &= \frac{1}{n} \sum_{i=1}^{n} \left|\frac{ y_i - \hat{y}_i }{y_i}\right| \times 100\% 
\end{align}
\end{subequations}
where $y_i$ and $\hat{y}_i$ are the $i$-th true and predicted outputs. $n$ is the number of points in the testing horizon.

\section{Simulation Results}
In this section, three case studies are carried out to show the effectiveness of the proposed framework. Case 1 shows the ablation study results. Case 2 compares the baseline models in section III-C and our proposed model. Case 3 shows the generalization capability, for which we add the feature weighting mechanism and error correction module to the Informer.

\subsection{Case 1: Ablation Study and Discussion}

An ablation study is conducted based on the NAU dataset. Table \ref{table:Abla} presents the MAE and MAPE results. The first row shows the performance of the backbone encoder-decoder based BiLSTM model. We then compare three different approaches of feature weighting: mutual information (MI) \cite{ghadimi2018two}, random forest (RF) \cite{xuan2021multi}, and our proposed feature weighting attention (FW). In addition, we evaluate the performance of the backbone model with two types of temporal attention mechanisms: single layer temporal attention (TA) and the proposed hierarchical temporal attention, which incorporates similar day information (SDA). Finally, we include the results for two error correction methods: the baseline ARIMA model (BL) \cite{duan2021short} and our proposed feature reinforced error correction model (EC). The results show that the proposed feature weighting and error correction outperform the existing methods. Each individual module improves the accuracy of the backbone model. Moreover, the combination of these modules further enhances the performance. Compared with the other competing alternatives, our proposed framework achieves a significant improvement in accuracy.

\begin{figure}[!htb]
\centering
  \includegraphics[width=0.5\textwidth]{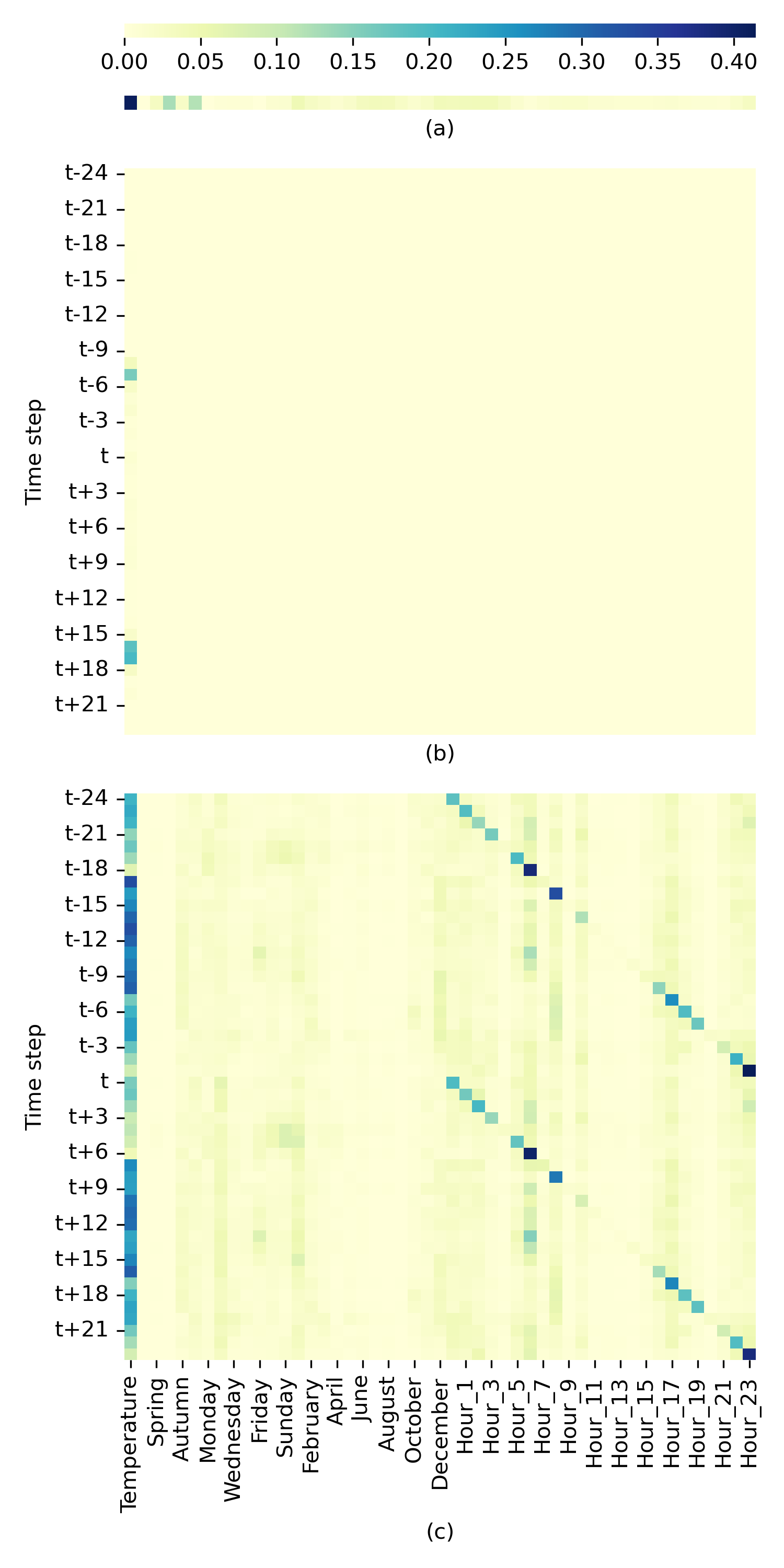}
  \caption{Two-day feature weight visualization for the NAU dataset with different approaches: (a) mutual information, (b) random forest, and (c) the proposed feature weighting attention.}\label{fig:feature_Utility}
\end{figure}

The interpretability of the performance improvements can be visualized in Fig \ref{fig:feature_Utility}. First, the proposed weighting attention can identify feature importance in the time domain. Fig \ref{fig:feature_Utility}(c) shows that our method adds time-varying weights on different features while mutual information approach exerts time-invariant weights shown in Fig \ref{fig:feature_Utility}(a).
Second, our method puts more accurate weights on each feature compared with RF. Our weight assignments are sparser, with higher weights on temperature and hour [cf. Fig. \ref{fig:feature_Utility}(c)]. In contrast, RF yields  similar features weights at approximately 0.02 [cf. Fig. \ref{fig:feature_Utility}(b)]. Third, our approach shows a good response to the input changes while RF is ignorant of different input data. Finally, the clear pattern of feature weights in Fig. \ref{fig:feature_Utility}(c) shows that the temperature from 9 AM to 4 PM is a more critical factor, which reflects the reality.

\begin{table}[tb]
\caption{NAU dataset: Ablation study for the proposed framework. Acronyms: MI (mutual information feature weight), RF (random forest feature weight), FW (feature weighting attention), TA (temporal attention), SDA (similar day attention), BL (baseline ARIMA error correction) and EC (the proposed error correction).}
\label{table:Abla}
\centering
\renewcommand{\arraystretch}{1.3}
\resizebox{\columnwidth}{!}
{%
\begin{tabular}{ccccccccc}
\hline
MI \cite{ghadimi2018two} & RF \cite{xuan2021multi} & FW & TA & SDA & BL \cite{duan2021short} & EC & MAE   & MAPE (\%) \\ \hline
-     & -     & -   & -  & -   & -        & -     & 89.20  & 3.93      \\
\checkmark     & -     & -   & -  & -   & -        & -     & 59.93 & 2.57      \\
-     & \checkmark     & -   & -  & -   & -        & -     & 63.65 & 2.76      \\
-     & -     & \checkmark   & -  & -   & -        & -     & 56.65 & 2.45      \\
-     & -     & -   & \checkmark  & -   & -        & -     & 69.10  & 2.96      \\
-     & -     & -   & \checkmark  & \checkmark   & -        & -     & 64.29 & 2.78      \\
-     & -     & \checkmark   & \checkmark  & \checkmark   & -        & -     & 48.96 & 2.15      \\
-     & -     & \checkmark   & \checkmark  & \checkmark   & \checkmark        & -     & 46.36 & 2.09      \\
-     & -     & \checkmark   & \checkmark  & \checkmark   & -        & \checkmark     & \textbf{45.70}  & \textbf{2.00}         \\ \hline
\end{tabular}%
}
\end{table}

\subsection{Case 2: Load Forecasting Model Comparison}

Besides our proposed model (FW+TA+SDA in Table \ref{table:Abla}) with LSTM (PM-LSTM) and GRU (PM-GRU) implementations, we test eight baseline models. To have a fair comparison, the error correction module is deactivated because it is not directly applicable to those classic machine learning algorithms such as SVR, RF and GBM. Table \ref{table:ISO_result} and Table \ref{table:Utility_result} show the forecasting error for the two datasets.

\begin{figure*}[!htb]
\centering
  \includegraphics[width=0.95\textwidth]{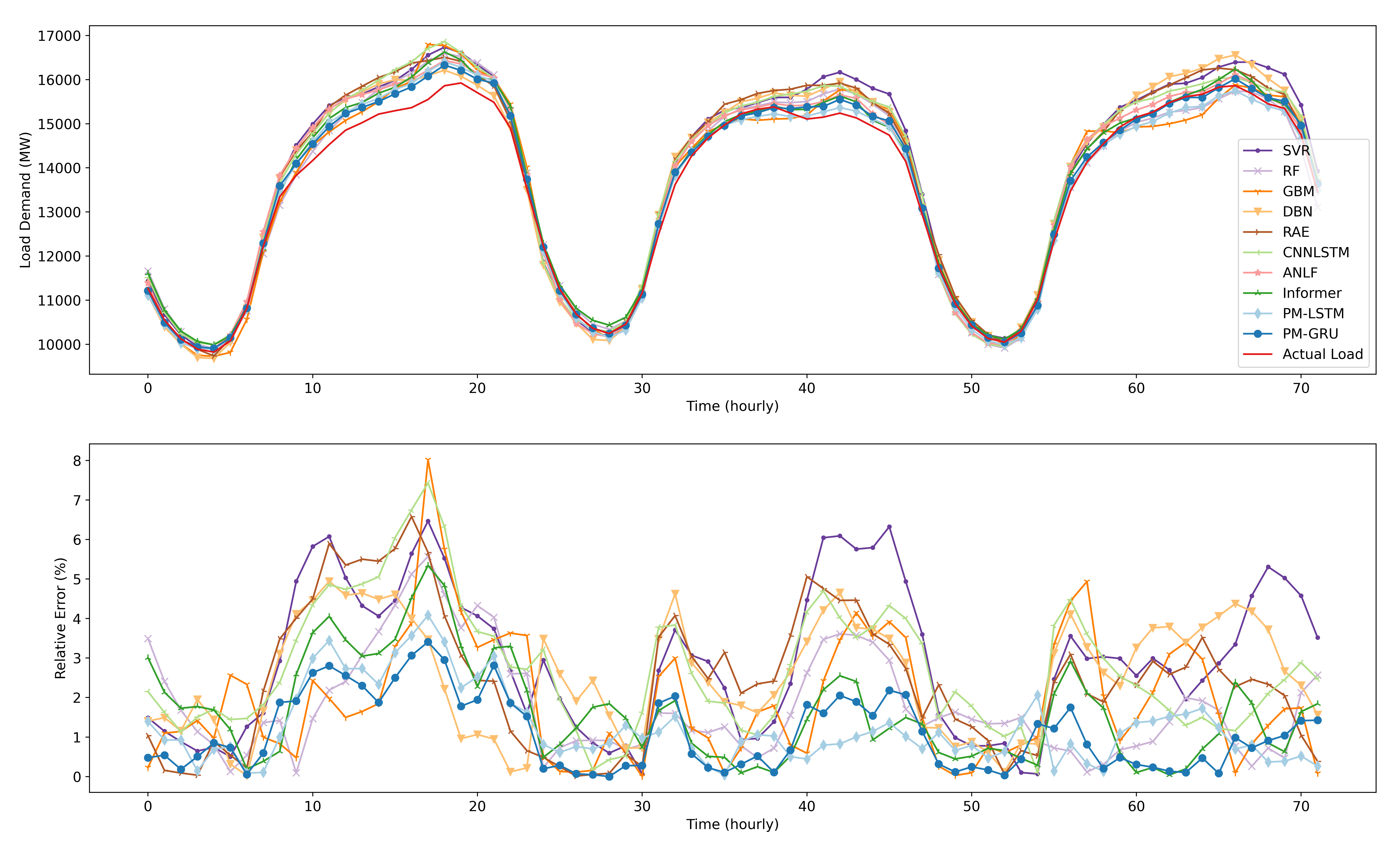}
  \caption{Forecasting curve and relative error for the ISO-NE dataset (3 days).}\label{fig:ISO_3day}
  \vspace{-0.5cm}
\end{figure*}

\begin{figure*}[!htb]
\centering
  \includegraphics[width=0.95\textwidth]{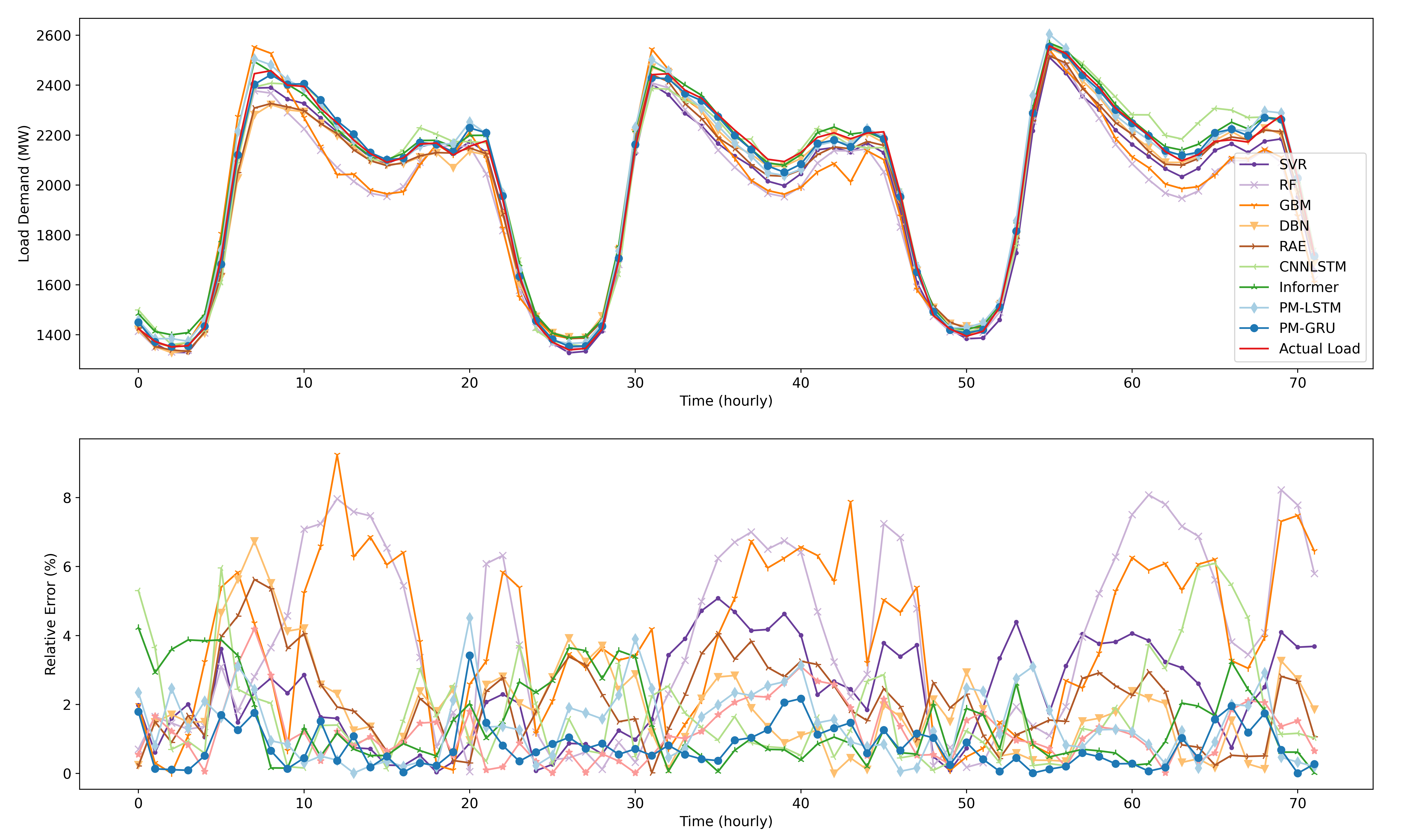}
  \caption{Forecasting curve and relative error for the NAU dataset (3 days).}\label{fig:Utility_3day}
  \vspace{-0.5cm}
\end{figure*}

Among the three classic machine learning methods, GBM performs the best for the ISO-NE dataset, while SVR stands out for the NAU dataset. DBN-based models attain results that are comparable to CNN-LSTM for both datasets. Notably, our proposed models PM-LSTM/GRU consistently outperform all the other models for both datasets with the smallest forecasting errors. Moreover, comparing the ANLF model with the proposed ones, extracting the feature weighting layer further improves the accuracy and increases the model's generalization capability. Overall, these findings highlight the effectiveness of our proposed approach.
The detailed forecasting performance over three days are given in Fig.~\ref{fig:ISO_3day} and Fig.~\ref{fig:Utility_3day}, where the relative error (RE) between the forecast value $\hat{y}$ and the true value $y$ is defined as
\begin{align}
\mathrm{RE}=\frac{|y-\hat{y}|}{y}\times 100\%.
\end{align}

\begin{table*}[!htb]
\caption{Forecasting errors over the year 2019 for the ISO-NE dataset.}
\label{table:ISO_result}
\centering
\renewcommand{\arraystretch}{1.3}
\resizebox{2\columnwidth}{!}
{%
\begin{tabular}{lcccccccccc}
\hline
Model                           & SVR \cite{ceperic2013strategy}    & RF \cite{dudek2015short}     & GBM \cite{taieb2014gradient} &DBN\cite{dedinec2016deep} & RAE\cite{khodayar2017rough} & CNN-LSTM \cite{rafi2021short}  & ANLF\cite{xiong2021short}   & Informer\cite{zhou2021informer}   & PM-LSTM & PM-GRU \\ \hline
{MAE}        & 317.49 & 393.58 & 277.54 &326.45 & 308.36  &309.06   & 258.70  & 256.89   & \textbf{229.47}  & \textbf{231.84} \\
{MAPE (\%)}  & 2.33   & 2.87   & 2.00   &2.38 & 2.22  &2.27   & 1.88   & 1.89    & \textbf{1.66}    & \textbf{1.67}  \\\hline
\end{tabular}%
}
\end{table*}

\begin{table*}[!htb]
\caption{Forecasting errors over the year 1991 for the NAU dataset.}
\label{table:Utility_result}
\centering
\renewcommand{\arraystretch}{1.2}
\resizebox{2\columnwidth}{!}
{%
\begin{tabular}{lcccccccccc}
\hline
Model      & SVR \cite{ceperic2013strategy}   & RF \cite{dudek2015short}     & GBM \cite{taieb2014gradient} &DBN\cite{dedinec2016deep} &RAE\cite{khodayar2017rough} &CNN-LSTM \cite{rafi2021short}  &  ANLF\cite{xiong2021short}  & Informer\cite{zhou2021informer}  & PM-LSTM & PM-GRU \\ \hline
MAE        & 67.97 & 99.91  & 83.05 &68.33 &63.55 &61.85    & 58.65 & 57.13     & \textbf{48.96}   & \textbf{46.42}  \\
MAPE (\%)  & 2.98  & 4.40    & 3.61 &2.99 &2.79   &2.73   & 2.58  & 2.49      & \textbf{2.15}    & \textbf{2.03}   \\\hline
\end{tabular}%
}
\end{table*}

\subsection{Case 3: Generalization Capability}

\begin{figure}[!htb]
\centering
  \includegraphics[width=0.48\textwidth]{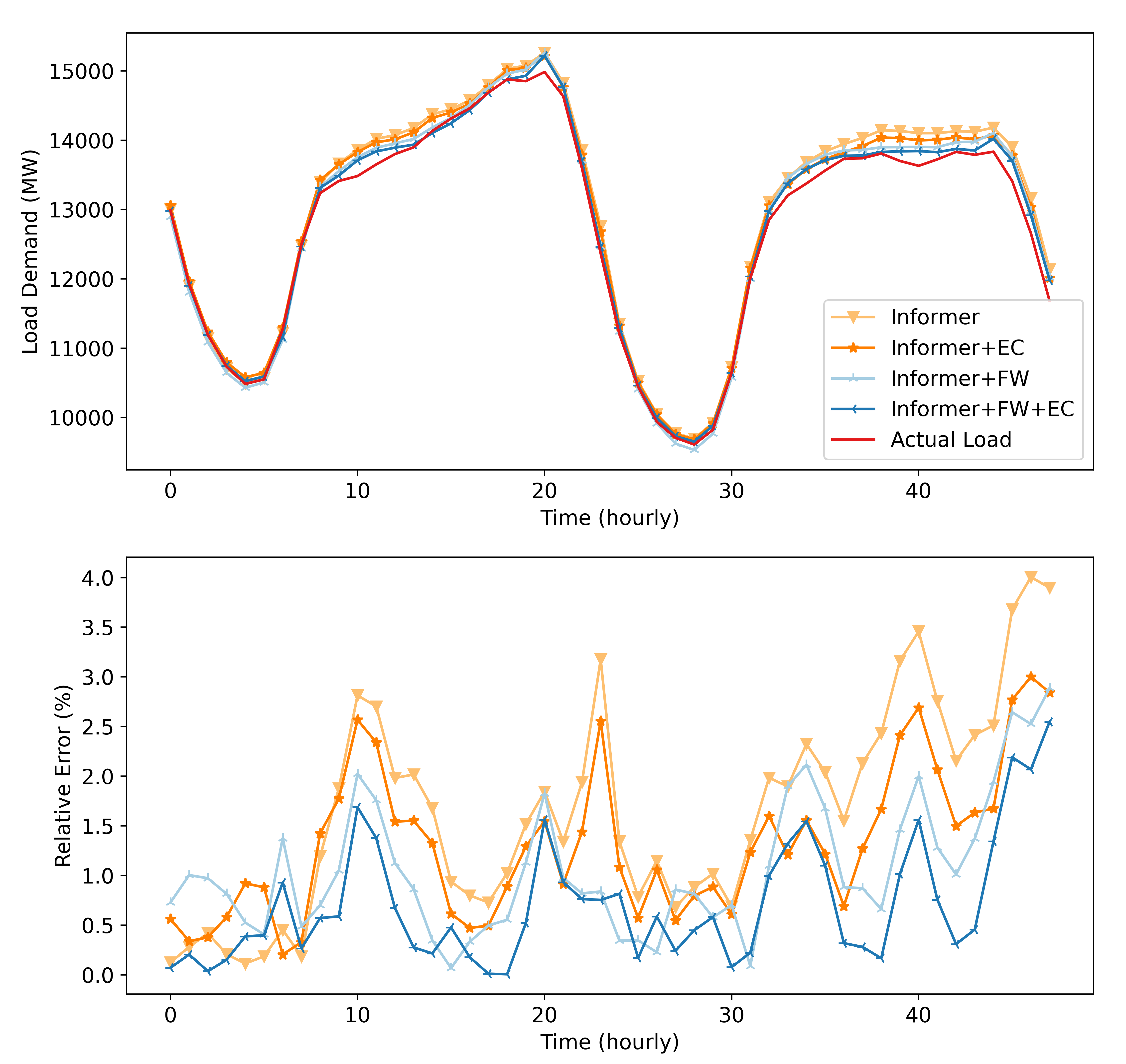}
  \caption{ISO-NE dataset: The forecasting curves and relative errors for Informer, Informer with error correction, Informer with feature weighting, and Informer with both feature weighting and error correction (for 2 days).}
  \label{fig:ISO_Informer_2_day_5592}
\vspace{-0.3cm}
\end{figure}

To further show the generalization capability of the proposed framework, we apply both the feature weighting and error correction to the transformer-based Informer. The result is reported in Table \ref{table:2} for the ISO-NE dataset. We compare the model itself with the model having feature weighting and/or error correction. The forecasting curves and relative errors are shown in Fig.~\ref{fig:ISO_Informer_2_day_5592}. 
By the ablation study, the model with our proposed feature weighting and error correction mechanisms performs the best. This verifies the merit of integrating the feature weighting to provide more informative features and error correction to further improve the accuracy. 

\begin{table}[tb]
\centering
\caption{ISO-NE dataset: Ablation study for Informer with feature reinforced error correction (EC) and/or feature weighting attention (FW).}
\label{table:2}
\renewcommand{\arraystretch}{1.3}{%
\begin{tabular}{lcccc}
\hline

FW & EC   & MAE    & MAPE (\%) \\ \hline
-   & -    & 256.89 & 1.89      \\
\checkmark&-    & 249.17 & 1.81      \\
- & \checkmark    & 239.35 & 1.76      \\
\checkmark & \checkmark  & \textbf{239.23} & \textbf{1.74}      \\ \hline
\end{tabular}%
}
\end{table}

\subsection{Case 4: Computational complexity}

In this section, we provide the big-$\mathcal{O}$ computational complexity analysis for the proposed framework. The feature weighting module has a computational complexity of $\mathcal{O}(nd^{fw}_h)$. The load forecasting model consists of the encoder and decoder BiLSTM whose complexity is $\mathcal{O}(T_h h_s (h_s + n))$ and $\mathcal{O}(T_f h_s( h_s + n))$, respectively \cite{rotman2021shuffling}. The hierarchical temporal attention that forms an additional input to the decoder BiLSTM has a complexity of $\mathcal{O}(T_h d^{att}_h + d^{att}_h(h_s+n))$. The output layer is in the $\mathcal{O}(d^o_hh_s)$. Considering all these components, the overall complexity of the load forecasting model is $\mathcal{O}(T_h h_s (h_s + n) + T_f h_s( h_s + n + T_h d^{att}_h + d^{att}_h(hs+n)) + d^o_hh_s)$. If $d^{fw}_h = d^{att}_h = d^{o}_h =: d_h$, then we have the overall computational complexity 
$\mathcal{O}(T_h h_s (h_s + n) + T_fh_sd_h (T_h + h_s + n))$.

\section{Conclusion}

This paper develops a unifying deep learning framework for multi-horizon STLF. Three interactive modules are developed with high generalization capability, which includes the feature weighting mechanism, STLF model and error correction module. In the proposed framework, the feature weighting mechanism is designed to provide informative input features for both historical and future time horizons. The STLF model with a hierarchical temporal attention layer decodes the next-day load with the future input features and similar temporal information. The hierarchical temporal attention layer provides a natural way to incorporate similar day information. 
In addition, the error correction module is developed based on transfer learning. It can reuse the learned hidden feature extraction to reduce the training cost. The modular design of our framework facilitates customization and independent modification. The extensive simulation results tested on the two datasets corroborate the merits of our framework. The codes of this work are available at \url{https://github.com/jxiong22/STLF_framework}

\nocite{*}
\bibliographystyle{IEEEtran}
\bibliography{refs}

\vspace{-20pt}
\begin{IEEEbiography} [{\includegraphics[width=1in,height=1.25in,clip,keepaspectratio]{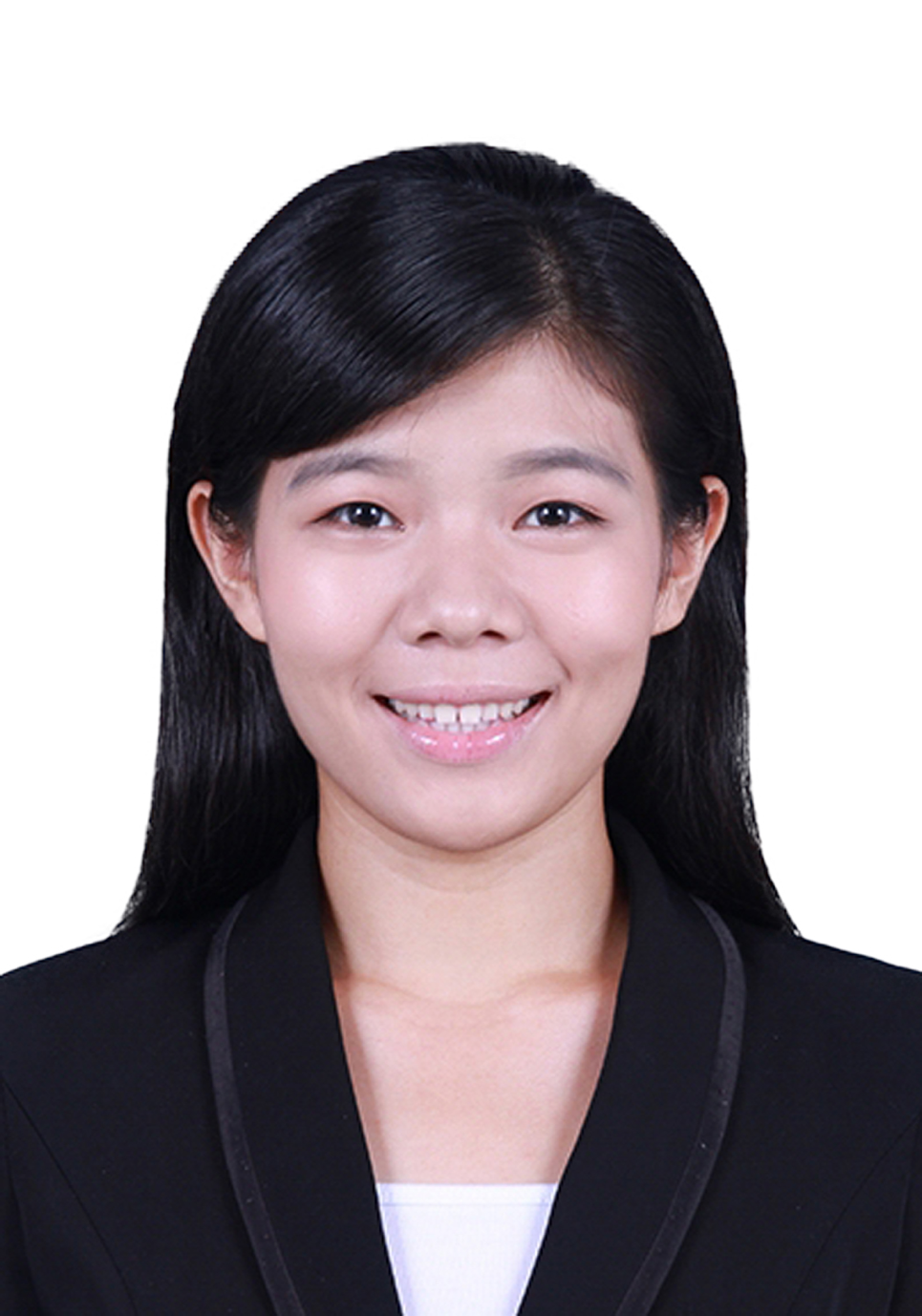}}]
{Jing Xiong} received  the B.S. and M.S. degrees in the Control and Computer Engineering department from North China Electric Power University. 
She is currently a Ph.D. candidate in the Electrical and Computer Engineering department at the University of California, Santa Cruz. Her research interests lie in big data analytics for smart power grids, load forecasting and monitoring, and event identification.
\end{IEEEbiography}

\vspace{-20pt}
\begin{IEEEbiography} [{\includegraphics[width=1in,height=1.25in,clip,keepaspectratio]{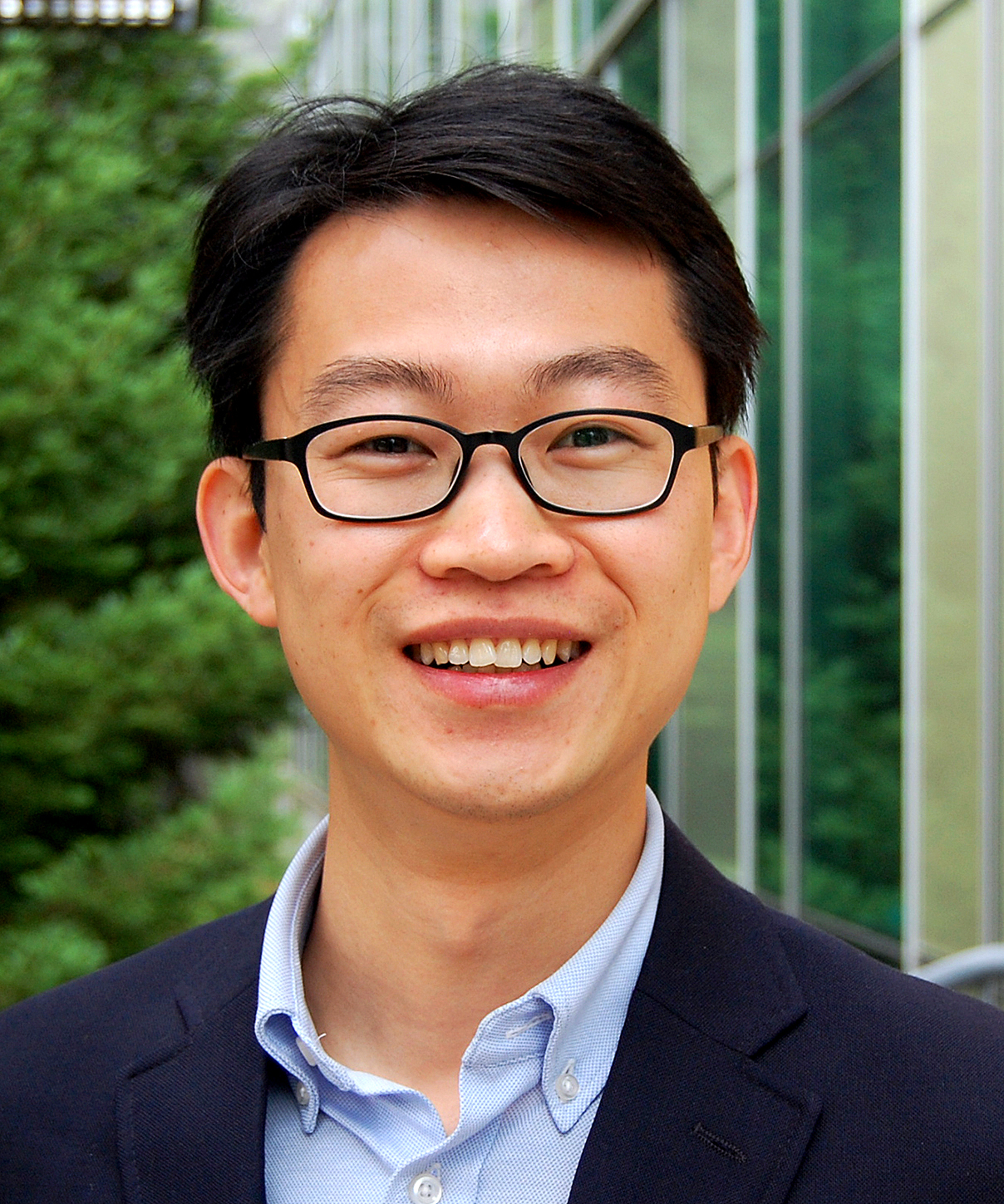}}]
{Yu Zhang} (M'15) is an Assistant Professor in the ECE Department of UC Santa Cruz. Prior to joining UCSC, he was a postdoc at UC Berkeley and Lawrence Berkeley National Laboratory. He received the Ph.D. degree in Electrical and Computer Engineering from the University of Minnesota. Dr. Zhang's research interests span the broad areas of cyber-physical systems, smart power grids, optimization theory, machine learning and big data analytics. Dr. Zhang received the Hellman Fellowship in 2019. He was the co-recipient of the Early Career Best Paper Award given by the Energy, Natural Resources, and the Environment (ENRE) section of the Institute of Operations Research and the Management Sciences (INFORMS) in 2021.
\end{IEEEbiography}
\EOD
\end{document}